\documentclass[10pt, conference,dvipsnames]{IEEEtran}

\usepackage{subfig}
\usepackage[utf8]{inputenc} 
\usepackage{graphicx,url}
\usepackage{booktabs}       
\graphicspath{ {./images/} }
\usepackage{multirow}
\usepackage{hhline}
\usepackage{graphicx}
\usepackage{amssymb,amsmath}
\usepackage{color,soul}
\usepackage[dvipsnames]{xcolor}
\usepackage{bbm}
\usepackage{float}
\makeatletter
\newif\if@restonecol
\makeatother

\usepackage[linesnumbered,ruled,vlined]{algorithm2e}
\usepackage{algpseudocode}
\usepackage{amsmath}
\usepackage{fancyhdr} 
\usepackage{tabularx}
\usepackage{comment} 

\usepackage{tikz}
\usetikzlibrary{shapes.geometric,backgrounds,patterns, trees}
\usetikzlibrary{3d,decorations.text,shapes.arrows,positioning,fit,backgrounds}
\usetikzlibrary{positioning, decorations.pathmorphing, shapes}
\usetikzlibrary{shapes.geometric,backgrounds,patterns, trees}
\usetikzlibrary{arrows.meta,
                bending,
                intersections,
                quotes,
                shapes.geometric}
              \usetikzlibrary{automata, positioning}
\usetikzlibrary{shapes,arrows}
\usetikzlibrary{arrows.meta}
\usetikzlibrary{positioning}
\tikzset{set/.style={draw,circle,inner sep=0pt,align=center}}
\usetikzlibrary{automata, positioning}
  \usetikzlibrary{shapes,shadows}
  \tikzstyle{abstractbox} = [draw=black, fill=white, rectangle,
  inner sep=10pt, style=rounded corners, drop shadow={fill=black,
  opacity=1}]
\tikzstyle{abstracttitle} =[fill=white]
\usetikzlibrary{calc,positioning,shapes.geometric}
\usetikzlibrary{arrows.meta,arrows}
\DeclareMathOperator*{\argmax}{arg\,max}

\renewcommand\thetable{\arabic{table}}
\usepackage{etoolbox}
\makeatletter
\patchcmd{\@makecaption}
  {\scshape}
  {}
  {}
  {}
\makeatother
\begin{document}

\title{Finding Effective Security Strategies through Reinforcement Learning and Self-Play
}
\author{\IEEEauthorblockN{Kim Hammar \IEEEauthorrefmark{2}\IEEEauthorrefmark{3} and Rolf Stadler\IEEEauthorrefmark{2}\IEEEauthorrefmark{3}}

 \IEEEauthorblockA{\IEEEauthorrefmark{2}
Division of Network and Systems Engineering, KTH Royal Institute of Technology, Sweden
 }
 \IEEEauthorblockA{\IEEEauthorrefmark{3} KTH Center for Cyber Defense and Information Security, Sweden \\
  \newline
Email: \{kimham, stadler\}@kth.se
\\
\today
}
}
\maketitle
\begin{abstract}
We present a method to automatically find security strategies for the use case of intrusion prevention. Following this method, we model the interaction between an attacker and a defender as a Markov game and let attack and defense strategies evolve through reinforcement learning and self-play without human intervention. Using a simple infrastructure configuration, we demonstrate that effective security strategies can emerge from self-play. This shows that self-play, which has been applied in other domains with great success, can be effective in the context of network security. Inspection of the converged policies show that the emerged policies reflect common-sense knowledge and are similar to strategies of humans. Moreover, we address known challenges of reinforcement learning in this domain and present an approach that uses function approximation, an opponent pool, and an autoregressive policy representation. Through evaluations we show that our method is superior to two baseline methods but that policy convergence in self-play remains a challenge.
\end{abstract}
\section{Introduction}
An organization's security strategy has traditionally been defined, implemented, and updated by domain experts. Although this approach can provide basic security for an organization's communication and computing infrastructure, a growing concern is that update cycles become shorter and security requirements change faster \cite{maloof}. Two factors contribute to this development: on the one hand, infrastructures keep changing due to functional upgrades and innovation, and, on the other hand, attacks increase in sophistication. As a response, significant efforts are made to \textit{automate security processes and functions}, including automated threat modeling \cite{honeypot_game, mal_pontus}, game-theoretic approaches \cite{nework_security_alpcan, game_t_sec_survey}, evolutionary methods \cite{armsrace_malware,evolvable_malware}, and learning-based solutions \cite{closed_world,deep_rl_cyber_sec}.

To automate the response to a cyber attack in a changing environment, we view the interaction between an attacker and a defender as a game, and model the \textit{evolution of attack and defense strategies} with automatic self-play. To avoid the need for domain knowledge to be coded into agent strategies, we study methods where the strategies evolve without human intervention from random configurations. In particular, we use self-play, a setting where both the attacker and the defender learn by interacting with each other, as a technique to automatically harden the defense.

Formally, we model the interaction between an attacker and a defender as a Markov game. We then use simulations of self-play where autonomous agents interact and continuously update their strategies based on experience from previously played games. To automatically update strategies in the game, several methods can be used, including computational game theory, evolutionary algorithms, and reinforcement learning. In this work, we choose reinforcement learning, which allows us to use simulations of self-play where both agents evolve their strategies simultaneously.

In pursuing this approach, we face several challenges. First, from a game-theoretic perspective, the game between an attacker and a defender is one with partial observability. That is, a defender can typically not observe the state and location of an attack, and an attacker is generally not aware of the full state of a defender's infrastructure. Second, due to the scale and complexity of computer infrastructures, the games that we consider have large state and action spaces, which make them hard to analyze theoretically. Third, as both the attacker and the defender learn simultaneously during self-play, the reinforcement learning environment becomes non-stationary \cite{bowling_marl_1}, which is a known challenge for reinforcement learning and often results in agent strategies that do not converge \cite{local_opt_1,deepmind_2,dota_openai_1}.

We develop our method around the use case of \emph{intrusion prevention}. This use case considers a defender that owns an infrastructure that consists of a set of connected components (e.g. a communication network), and an attacker that seeks to intrude on the infrastructure. The defender's objective is to protect the infrastructure from intrusions by monitoring the network and patching vulnerabilities. Conversely, the attacker's goal is to compromise the infrastructure and gain access to a critical component. To achieve this, the attacker must explore the infrastructure through reconnaissance and attack components along the path to the critical component. In this work, we study a relatively small and simple infrastructure of the defender, which allows us to easier compare security strategies and focus on the principles of our method. 

With this paper, we make the following contributions. First, we show, using a small infrastructure, that reinforcement learning and self-play can be used to automatically discover effective strategies for intrusion prevention. Although self-play has been used successfully in other domains, such as board games \cite{deepmind_2} and video games \cite{deepmind_5,dota_openai_1}, our work is, to the best of our knowledge, the first study of self-play in the context of network and systems security. Second, we present a reinforcement learning method to address the challenges that appear in the context of our use case and show through evaluations that it outperforms two baselines. Specifically, to tackle the problem of the high dimensional state space, we use function approximation implemented with neural networks. To address the non-stationary reinforcement learning environment, we use an opponent-pool technique, and to deal with the large action space, we propose an autoregressive policy representation.

The remainder of this paper is structured as follows. We first cover the necessary background information on partially observed Markov decision processes, reinforcement learning, and Markov games. Then we introduce a game model for intrusion prevention and present our method together with experimental results. Lastly, we describe how our work relates to prior research, present our conclusions, and provide directions for future work.

\section{POMDPs, Reinforcement Learning and Markov Games}
This section contains background information on Markov decision processes, reinforcement learning, and Markov games.
\subsection{Markov Decision Processes}
A Markov Decision Process (MDP) is defined by a six-tuple $\mathcal{M} = \langle \mathcal{S}, \mathcal{A}, \mathcal{P}_{ss^{\prime}}, \mathcal{R}_{ss^{\prime}}^a, \gamma, \rho_0 \rangle$ \cite{bellman1957markovian}. $\mathcal{S}$ denotes the set of states and $\mathcal{A}$ denotes the set of actions. $\mathcal{P}^{a}_{ss^{\prime}}$ refers to the transition probability of moving from state $s$ to state $s^{\prime}$ when taking action $a$ (Eq. \ref{eq:mdp_prob_1}), which has the Markov property (Eq. \ref{eq:markov_prop_1}). Similarly, $\mathcal{R}^a_{ss^{\prime}} \in \mathbb{R}$ is the expected reward when taking action $a$ in state $s$ to transition to state $s^{\prime}$ (Eq. \ref{eq:mdp_reward_1}). Finally, $\gamma \in \left(0,1\right]$ is the discount factor and $\rho_0 : \mathcal{S} \mapsto [0,1]$ is the initial state distribution.
\begin{align}
  \mathcal{P}^{a}_{ss^{\prime}} &= \mathbb{P}\left[s_{t+1} = s^{\prime}| s_t = s, a_t = a\right] \label{eq:mdp_prob_1}\\
\mathbb{P}\left[s_{t+1}|s_t\right] &= \mathbb{P}\left[s_{t+1}| s_1, \hdots, s_t\right]\label{eq:markov_prop_1}\\
\mathcal{R}^a_{ss^{\prime}} &= \mathbb{E}\left[r_{t+1}| a_t = a,  s_t = s, s_{t+1} = s^{\prime}\right] \label{eq:mdp_reward_1}
\end{align}
A Partially Observed Markov Decision Process (POMDP) is an extension of an MDP \cite{howard_mdps,pomdps}. The difference compared with an MDP is that, in a POMDP, the agent does not directly observe the states $s \in \mathcal{S}$. Instead, the agent observes observations $o \in \mathcal{O}$, which depend on the state $s$ as defined by an observation function $\mathcal{Z}$. Specifically, a POMDP is defined by an eight-tuple $\mathcal{M}_{\mathcal{P}} = \langle \mathcal{S}, \mathcal{A}, \mathcal{P}^{a}_{ss^{\prime}}, \mathcal{R}^a_{ss^{\prime}},\gamma, \rho_0, \mathcal{O}, \mathcal{Z}\rangle$. The first six elements define an MDP. $\mathcal{O}$ denotes the set of observations and $\mathcal{Z} = \mathbb{P}[o|s]$ denotes the observation function, where $o \in \mathcal{O}$ and $s \in \mathcal{S}$.


\subsection{The Reinforcement Learning Problem}\label{sec:rl_prob2}
The reinforcement learning problem can be formulated as that of finding a policy $\pi^{*}$ that maximizes the expected cumulative reward of an MDP over a finite horizon $T$ \cite{rl_bible}:
\begin{align}
\pi^{*} &=\argmax_{\pi} \mathbb{E}\left[\sum_{t=0}^{T}\gamma^tr_{t+1}\right] \label{eq:rl_prob}
\end{align}
The Bellman equations \cite{bellman_eq} (Eq. \ref{eq:bellman_eq_31}-\ref{eq:bellman_eq_33}) relates the optimal policy $\pi^{*}$ to the states and the actions of the MDP, where $\pi^{*}(s) = \argmax_a Q^*(s,a)$.
\begin{align}
V^{*}(s) &= \displaystyle\max_a \mathbb{E}\left[r_{t+1} + \gamma V^{*}(s_{t+1}) | s_t = s, a_t = a\right]\label{eq:bellman_eq_31} \\
  Q^{*}(s,a) &= \mathbb{E}[r_{t+1} + \gamma \displaystyle\max_{a^{\prime}}Q^{*}(s_{t+1},a^{\prime}) | s_t = s, a_t = a] \label{eq:bellman_eq_33}
\end{align}
A reinforcement learning \textit{algorithm} is a procedure to compute or approximate $\pi^{*}$. Three main classes of such algorithms exist: value-based algorithms (e.g. Q-learning \cite{watkins_thesis}), policy-based algorithms (e.g. PPO \cite{ppo}), and model-based algorithms (e.g. Dyna-Q \cite{rl_bible}).
\subsection{Markov Games}
Markov games, also known as stochastic games \cite{Shapley1095}, provide a theoretical framework for \textit{multi-agent} reinforcement learning \cite{littman_marl_1}. A Markov game of $N$ agents is defined as a tuple $\mathcal{M}_G = \langle \mathcal{S}, \mathcal{A}_1, \hdots, \mathcal{A}_N,  \mathcal{T}, \mathcal{R}_{1}, \hdots \mathcal{R}_{N}, \gamma,  \rho_0 \rangle$. $S$ denotes the set of states and the list $\mathcal{A}_1, \hdots, \mathcal{A}_N$ denotes the sets of actions of agents $1, \hdots, N$. The transition function $\mathcal{T}$ is an operator on the set of states and the combined action space of all agents, $\mathcal{T}: \mathcal{S} \times \mathcal{A}_1 \times \hdots \times \mathcal{A}_N \mapsto \mathcal{S}$. Similarly, the functions $\mathcal{R}_1, \hdots \mathcal{R}_N$ define the rewards for agent $1, \hdots, N$ where $\mathcal{R}_{i} : \mathcal{S} \times \mathcal{A}_i \mapsto \mathbb{R}$. Moreover, $\gamma$ is the discount factor and $\rho_0: \mathcal{S} \mapsto [0,1]$ is the initial state distribution. Finally, in the partially observed setting of a Markov game, the model also includes sets of observations $\mathcal{O}_1, \hdots, \mathcal{O}_N$ as well as observation functions $\mathcal{Z}_1, \hdots, \mathcal{Z}_N$ \cite{marl_1}.
\section{Modeling Intrusion Prevention as a Markov Game}\label{sec:modeling}
In this section, we model the use case of intrusion prevention as a zero-sum Markov game that involves two agents, an attacker and a defender.

\subsection{Intrusion Prevention as a Game}\label{sec:modeling2}
\begin{figure}
  \centering
  \scalebox{0.67}{
    \input{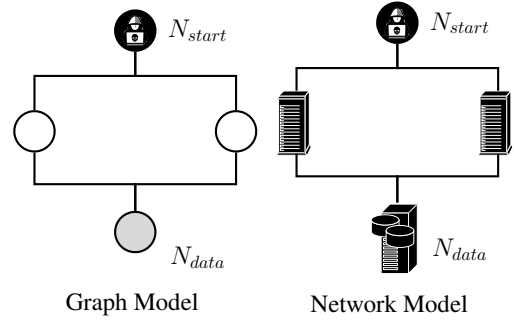}
    }
    \caption{Modeling intrusion prevention as a Markov game.}
    \label{fig:model_graph}
\end{figure}

The right side of Fig. \ref{fig:model_graph} shows the infrastructure underlying our use case. It is depicted as a graph that includes four network components. The component $N_{start}$ represents the attacker's computer and the remaining components represent the defender's infrastructure, where $N_{data}$ is the component that the attacker wants to compromise.

\begin{figure*}
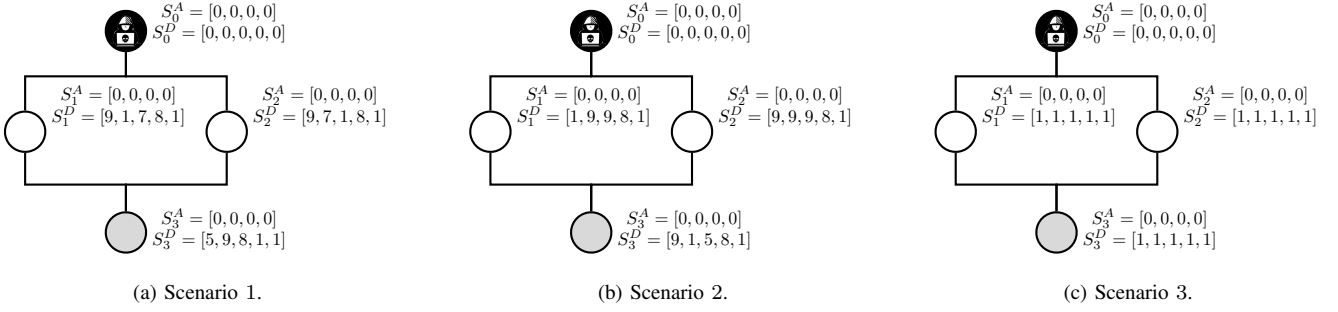

  \centering
  \subfloat[Scenario $1$.]{
  \scalebox{0.67}{
    \input{tikz/version_2.tex}
    }\label{fig:version_1}
  }\quad
  \hfill
  \subfloat[Scenario $2$.]{
        \scalebox{0.67}{
          \input{tikz/version_1.tex}
         }
         \label{fig:version_2}}
\quad\hfill
  \subfloat[Scenario $3$.]{
        \scalebox{0.67}{
          \input{tikz/version_3.tex}
         }
  \label{fig:version_3}}
\caption{Three scenarios of the intrusion prevention game. Scenarios $1$ and $2$ model infrastructures with strong defenses (first four defense attributes) but weak detection capabilities (fifth defense attribute). In scenario $1$, each node contains one vulnerability (a defense attribute with a value $\leq 1$), whereas in scenario $2$ only one of the intermediary nodes has a vulnerability (the left one). Scenario $3$ models an infrastructure with both weak defenses and weak detection capabilities.}\label{fig:versions}
\end{figure*}
To compromise the target component $N_{data}$, the attacker must explore the infrastructure through reconnaissance and compromise components on the path to $N_{data}$. At the same time, the defender monitors the network and increases its defenses to prevent the attacker from reaching $N_{data}$. In this adversarial process, both the attacker and the defender have a partial view of the network. At the beginning of the game, the attacker knows neither the topology of the infrastructure nor its vulnerabilities. In contrast, the defender has complete knowledge of the network's topology and the vulnerabilities of its components, but cannot observe the status of attacks.

The described adversarial process evolves as a round-based game. In each round, the attacker and the defender perform actions on components in the network, continuing until either the attacker wins the game by compromising $N_{data}$, or the defender wins the game by detecting the attacker.
\subsection{A Formal Game Model of Intrusion Prevention}
We extend prior work by others \cite{elderman} and model the network infrastructure as a graph $\mathcal{G} = \langle \mathcal{N}, \mathcal{E} \rangle$ (see Fig. \ref{fig:model_graph} left side). The nodes $\mathcal{N}$ of the graph represent the components of the infrastructure. Similarly, the edges $\mathcal{E}$ of the graph represent connectivity between components. The graph has two special nodes, $N_{start}$ and $N_{data}$, representing the attacker's starting position and target, respectively. Moreover, each node $N_k \in \mathcal{N}$ has an associated node state, $S_k = \langle S_k^{A}, S_k^{D} \rangle$. The node's defense status $S_k^{D}$ is only visible to the defender and consists of $m+1$ attributes, $S^{D}_{k,1} \hdots, S^{D}_{k,m+1}$. Likewise, the node's attack status $S^{A}_k$ is only visible to the attacker and consists of $m$ attributes, $S^{A}_{k,1}, \hdots S^{A}_{k,m}$. The attribute values are natural numbers in the range $0,\hdots,w$, where  $w > 0$.

The values of the attack attributes $S^A_{k,1}, \hdots, S^A_{k,m}$ represent the strength of $m$ different types of attacks against node $N_k$, e.g. denial of service attacks, cross-site scripting attacks, etc. Similarly, the values of the defense attributes $S^D_{k,1}, \hdots, S^D_{k,m}$ represent the strength of the node's defenses against the $m$ attack types and encodes the node's security mechanisms, such as firewalls and encryption functions. Additionally, each node $N_k$ has a special defense attribute $S^{D}_{k,m+1}$, whose value represents the node's capability to detect an attack.

As an example, installing an intrusion detection system\textemdash a monitoring mechanism\textemdash at node $N_k$ would correspond to an increase of the detection value $S^{D}_{k,m+1}$ in the model. Correspondingly, installing a rate-limiting mechanism\textemdash a security mechanism against denial of service attacks\textemdash at node $N_k$ would correspond to an increase of the defense attribute $S^D_{k,j}$, where $j$ is the attack type for denial of service attacks in the model.

The attacker can perform two types of actions on a visible node $N_k$: a reconnaissance action, which renders the defense status $S_k^D$ visible to the attacker, or an attack of type $j\in 1 \hdots m$, which increases the attack value $S^A_{k,j}$ by $1$. If the attack value exceeds the node's defense value, i.e. $S^A_{k,j} > S^D_{k,j}$, the attacker has compromised the node and the node's neighbors become visible to the attacker. Conversely, if the attacker performs an attack that does not compromise a node, the attack is detected by the defender with probability $p = \frac{S^{D}_{k,m+1}}{w+1}$, defined by the node's detection capability $S^{D}_{k,m+1}$.

Just like the attacker, the defender can perform two types of actions on node $N_k$: either a monitoring action, that improves the detection capability of the node and increments the detection value $S^{D}_{k,m+1}$, or a defensive action, that improves the defense against attacks of type $j \in 1 \hdots m$ and increments the defense value $S^{D}_{k,j}$.

The actions are performed on a round-by-round basis in the game. In each round, the attacker and the defender perform one action each, which brings the system into a new state. The game ends either when the attacker compromises the target node, $N_{data}$ (attacker wins), or when the attacker is detected (defender wins). The winner of a game is rewarded with a utility of $+1$ whereas the opponent receives a utility of $-1$.

The game evolves as a stochastic game with the Markov property, $\mathbb{P}\left[s_{t+1}|s_t\right] = \mathbb{P}\left[s_{t+1}| s_1, \hdots, s_t\right]$, which follows trivially from the game dynamics defined above. The size of the state space $\mathcal{S}$ of the Markov game is $(w+1)^{|\mathcal{N}|\cdot m \cdot (m+1)}$. Finally, the size of the action spaces $\mathcal{A}_1$ and $\mathcal{A}_2$ for both the attacker and the defender is $|\mathcal{N}|\cdot (m + 1)$.



\subsection{Scenarios for the Intrusion Prevention Game}\label{sec:scenarios}
The Markov game defined above is parameterized by the graph's topology, the number of attack types $m$, the maximal attribute value $w$, and the initial node states $S_{1}, \hdots, S_{|\mathcal{N}|}$. In the remainder of the paper we focus on three specific scenarios that are depicted in Fig. \ref{fig:versions}. We define the number of attack types to be four ($m = 4$), the maximal attribute value $w = 9$, and initialize all attack states to zero: $S^{A}_{k,i} = 0$ for $k \in 1 \hdots |\mathcal{N}|$ and $i \in 1 \hdots m$. The three scenarios differ in how their defense states $S^{D}_{k,i}$ for $k \in 1 \hdots |\mathcal{N}|$ and $i \in 1 \hdots m+1$ are initialized, as is illustrated in Fig. \ref{fig:versions}. Specifically, scenario $1$ and $2$ model infrastructures with strong defenses but weak detection capabilities whereas scenario $3$ models an infrastructure with both weak defenses and weak detection capabilities.

\section{Finding Security Strategies through Reinforcement Learning}\label{sec:rl_method}
This section describes our method for finding strategies for the attacker and the defender in the game defined in Section \ref{sec:modeling}. Specifically, we describe how self-play in combination with reinforcement learning can be used to learn strategies without prior knowledge. We also discuss our approach to address the difficulties associated with the high-dimensional state and action spaces, as well as the challenges that come with a non-stationary environment.
\subsection{Learning Policies through Self-Play}
To simulate games using the model defined in Section \ref{sec:modeling}, actions for the attacker and the defender are sampled from policies\footnote{As we build on prior work in both reinforcement learning and game theory we use the terms ``strategy'' and ``policy'' interchangeably.}. In our approach, these policies are not defined by experts, but evolve through the process of self-play, which we implement as follows. First, we initialize both the attacker and the defender with random policies. Next, we run a series of simulations where the agents play against each other using their current policies. After a given number of games, we use the game outcomes and the game trajectories to update the policies using reinforcement learning. This process of playing games and updating the agents' policies continues until both policies sufficiently converge. Although self-play often converges in practice, as reported in \cite{td_gammon,deepmind_2,dota_openai_1}, no formal guarantees for policy convergence in self-play have been established.
\paragraph*{The Challenge of a Non-Stationary Environment}
One reason for the lack of convergence guarantees of self-play is that the environment is non-stationary, which is an inherent challenge for reinforcement learning \cite{bowling_marl_1,local_opt_1}. Specifically, when multiple agents learn simultaneously in self-play, the agents' policies are part of the environment. This means that for each agent, the environment changes when the opponent updates its policy. Hence, the environment is non-stationary. To improve the possibility of convergence in self-play, despite the non-stationary environment, different methods have been proposed in prior work \cite{deepmind_2, dota_openai_1}. We describe our approach to this issue in Section \ref{sec:opponent_pool}.





\subsection{Reinforcement Learning Algorithm}\label{sec:rl_algos}
To learn strategies for the attacker and the defender in the game, we use the well-known reinforcement learning algorithms REINFORCE \cite{williams_1} and PPO \cite{ppo}. Both of them implement the \textit{policy gradient} method for solving the reinforcement learning problem (Eq. \ref{eq:rl_prob}). This method can be summarized as follows. First, the policy $\pi$ is represented as a parameterized function $\pi_{\theta}$, where $\theta \in \mathbb{R}^{d}$. Second, an objective function $J(\theta)$ that estimates the expected reward for the policy $\pi_{\theta}$ is introduced. For example, the expected reward following policy $\pi_{\theta}$ can be estimated by sampling from the environment, e.g. $J(\theta) = \mathbb{E}_{o \sim \rho^{\pi_{\theta}},a\sim \pi_{\theta}}[R]$, where $\rho^{\pi_{\theta}}$ is the observation distribution and $R$ is the cumulative episode reward. Third, the optimization problem of maximizing $J(\theta)$ is solved using stochastic gradient ascent with a variant of the following gradient:
\begin{align}
  \nabla_{\theta} J(\theta) &= \mathbb{E}_{o \sim \rho^{\pi_{\theta}},a\sim \pi_{\theta}}\left[\underbrace{\nabla_{\theta}\log\pi_{\theta}(a|o)}_{\text{actor}} \underbrace{A^{\pi_{\theta}}(o,a)}_{\text{critic}}\right] \label{eq:pg_objective}
\end{align}
where $A^{\pi_{\theta}}(o,a) = Q^{\pi_{\theta}}(o,a) - V^{\pi_{\theta}}(o)$ is the advantage function, that gives an estimate of how advantageous the action $a$ is when following policy $\pi_{\theta}$ compared to the average action.




\subsection{Function Approximation}
The Markov game described in Section \ref{sec:modeling} has a high-dimensional state space where the number of states grows exponentially with the number of nodes $|\mathcal{N}|$ and attack types $m$. Specifically, as each node $N_k \in \mathcal{N}$ has $m$ attack attributes and $m+1$ defense attributes, whose values range from $0, \hdots, w$, the state space has size $|\mathcal{S}| = (w+1)^{|\mathcal{N}|\cdot m \cdot (m+1)}$. This means that tabular reinforcement learning methods that rely on enumerating the entire state space are impractical. Instead, we use a parameterized function $\pi_{\theta}$ as a more compact representation of the policy than a table. In particular, we implement $\pi_{\theta}$ with a deep neural network that has a fixed set of parameters $\theta \in \mathbb{R}^{d}$. The neural network takes as input an observation $o$, and outputs a conditional probability distribution $\pi_{\theta}(a|o)$ over all possible actions $a$. Moreover, the neural network uses an actor-critic architecture \cite{gae} and computes a second output that estimates the value function $V^{\pi_{\theta}}(o)$, which is used to estimate $A^{\pi_{\theta}}(o,a)$ in the gradient in Eq. \ref{eq:pg_objective}.
\begin{figure*}
  \centering
    \scalebox{0.45}{
      \includegraphics{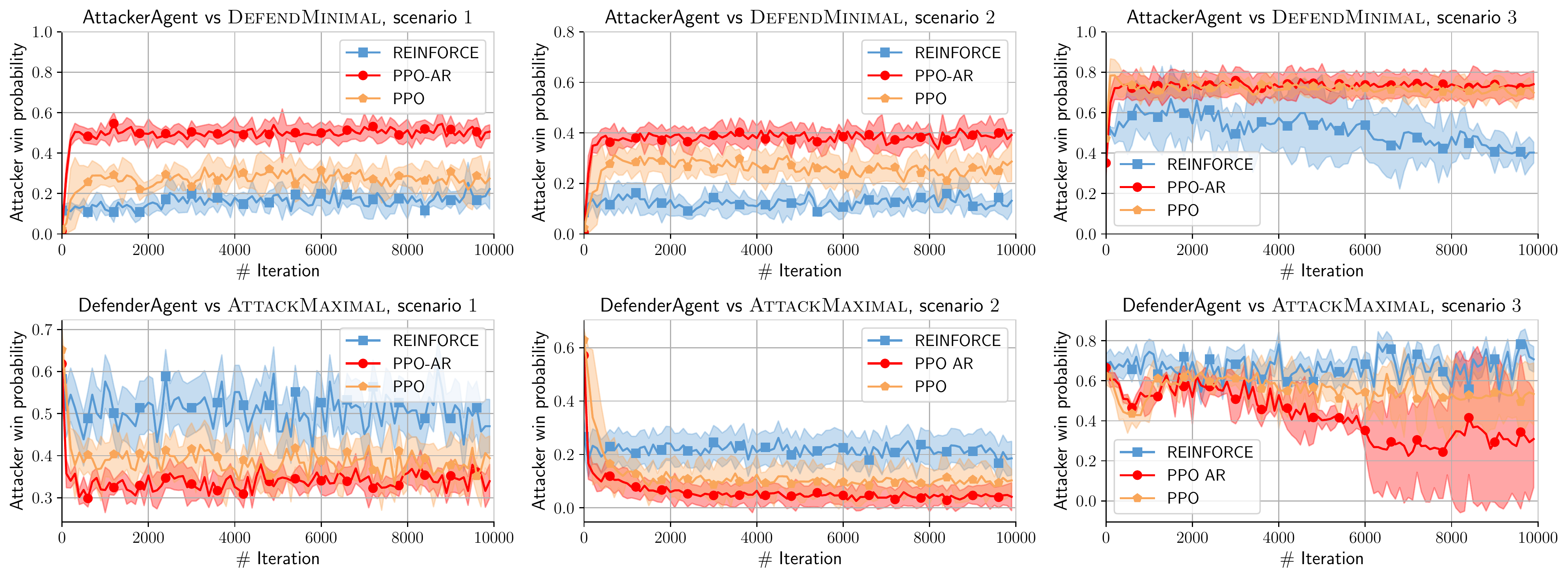}
    }
    \caption{Attacker win ratio against the number of training iterations; the top row shows the results from training the attacker against \textsc{DefendMinimal}; the bottom row shows the results from training the defender against \textsc{AttackMaximal}; the three columns represent the three scenarios; the curve labeled PPO-AR shows the mean values of our proposed method; the results are averages over five training runs with different random seeds; the shaded regions show the standard deviation.}
    \label{fig:best_response_learning}
\end{figure*}
\subsection{Autoregressive Policy Representation}\label{sec:auto_reg}
Using reinforcement learning in environments with large action spaces is challenging and often results in slow convergence of policies \cite{deepmind_5}. In our case, the action space of the Markov game from Section \ref{sec:modeling} grows linearly with the number of nodes $|\mathcal{N}|$ and the number of attack types $m$. To deal with this challenge, we leverage the structure of the problem domain to reduce the size of the action space. Specifically, we represent an action as a sequence of two sub-actions: (1) select the node $n$ to attack or defend; and (2) select the type of attack or defense $a$. As a consequence, we can decompose the policy $\pi(a, n|o)$ in two sub-policies, $\pi(a|n,o)$ and $\pi(n|o)$ based on the chain rule of probability (Eq. \ref{eq:chain_rule_p}). This reduces the size of the action space of both the attacker and the defender from the size $|\mathcal{N}|\cdot (m+1)$ to $|\mathcal{N}| + (m+1)$.
\begin{align}
\pi(a, n|o) = \pi(a|n,o) \cdot \pi(n|o)\label{eq:chain_rule_p}
\end{align}
We implement each sub-policy with a neural network. First, $\pi_{\theta}(n|o)$ takes an observation $o$ as input and decides the node $n$ to attack or defend. Next, the second neural network $\pi_{\phi}(a|n,o)$ focuses on the selected node $n$ to identify vulnerabilities that can be exploited or patched with an attack or defense of type $a$. Hence, the action is sampled in an \textit{autoregressive} manner from the two sub-policies.
\subsection{Opponent Pool}\label{sec:opponent_pool}
Although prior works have demonstrated that self-play can be a powerful learning method \cite{deepmind_2, dota_openai_1}, it is also well-known that self-play can suffer from training instability, which prevents the agents' policies from converging \cite{heinrich_1,deepmind_2,dota_openai_1}. Specifically, it can occur that an agent adjusts its policy to its current opponent to such a degree that the policy overfits and becomes ineffective against other policies. Our approach to mitigate this problem is to sample opponent policies from a \textit{pool} of policies during training \cite{deepmind_2, dota_openai_1}. The opponent pool increases the diversity of policies, and as a result, reduces the chance of overfitting. To populate the pool, we add the agents' current attack and defense policies periodically to the pool during training.
\section{Learning Effective Security Strategies}\label{sec:learning}
In this section, we evaluate the method presented in Section \ref{sec:rl_method} for finding security strategies for the use case of intrusion prevention. We train reinforcement learning agents using simulations of self-play
and evaluate our method with respect to convergence of agent policies. We also compare its performance with those of baseline reinforcement learning algorithms and inspect the learned policies. In sub-section \ref{sec:brpl} we train the attacker and the defender against static opponents, and in sub-section \ref{sec:dnmp} we train both the attacker and the defender simultaneously.

Our implementation of the game model and the agents' policies is open source and available at \texttt{\url{https://github.com/Limmen/gym-idsgame}}. The simulations have been conducted using a Tesla P100 GPU and hyperparameters are listed in Appendix \ref{appendix:hyperparameters}.
\subsection{Learning Policies Against a Static Opponent}\label{sec:brpl}
We first examine whether our method can discover effective attack and defense policies against a \textit{static} opponent policy. This means that we keep one policy fixed in self-play and that the other agent learns its policy against a static opponent. As one agent's policy is fixed, the environment for the other agent is stationary, which simplifies policy convergence.
\begin{figure*}
  \centering
  \scalebox{0.6}{
    \input{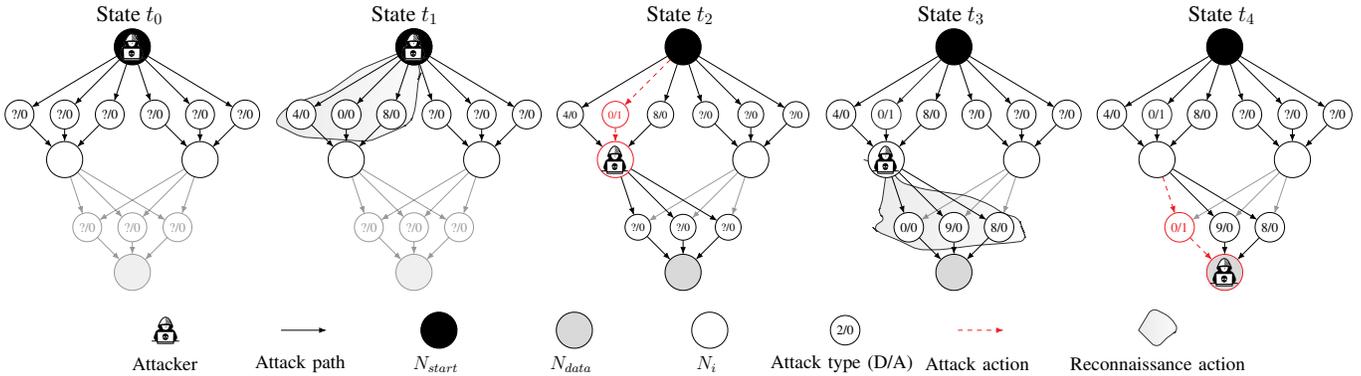}
    }
    \caption{An illustration of a learned attack strategy, evolving from left to right. The attacker first scans a neighboring node for vulnerabilities (low defense attributes) (state $t_1$). The attacker then exploits the found vulnerability (state $t_2$), compromises the node, and scans the target node $N_{data}$ (state $t_3$). Finally, the attacker completes the intrusion by attacking $N_{data}$ (state $t_4$).}
    \label{fig:idsgame_strategy}
  \end{figure*}
\subsubsection{Static Opponent Policies}
We investigate two cases of learning against a static opponent. In the first case, the attacker learns its policy by playing against a static defender policy called \textsc{DefendMinimal}. In the second case, the defender learns its policy by playing against a static attacker policy called \textsc{AttackMaximal}. \textsc{DefendMinimal} is a heuristic policy that updates the attribute with the minimal defense value across all nodes. Similarly, \textsc{AttackMaximal} is a heuristic policy that updates the attribute with the maximal attack value across all nodes that are visible to the attacker. If several minimal or maximal values exists, \textsc{DefendMinimal} and \textsc{AttackMaximal} picks the attribute to update at random.
\subsubsection{Simulation Setup}
To evaluate our method, we run simulations of the three scenarios described in Section \ref{sec:scenarios}. Before each simulation run, we initialize the model with a random permutation of the defense attributes to prevent overfitting (see Fig. \ref{fig:versions}). In addition to the three scenarios, we evaluate three algorithms: our proposed method, PPO-AR, that uses an autoregressive policy representation (see Section \ref{sec:auto_reg}), and two baseline methods, regular PPO and REINFORCE. Apart from the different algorithms and policy representations, all three methods use the same training setup and hyperparameters (Appendix \ref{appendix:hyperparameters}). For each scenario and algorithm, we run $10,000$ training iterations, each iteration consisting of $2,000$ game rounds. After each training iteration, the non-static agent updates its policy. Moreover, to measure stability we run each simulation five times with different random seeds, where a single simulation run accounts for $4$ hours of training time on a P100 GPU. The average results are shown in Fig. \ref{fig:best_response_learning}.


\subsubsection{Analysis of the results in Fig. \ref{fig:best_response_learning}}
The top row of Fig. \ref{fig:best_response_learning} shows the results from training the attacker against the static defender, and the bottom row shows the results from training the defender against the static attacker. The three columns represent the three scenarios. The x-axes denote the training iterations and the y-axes denote the attacker's empirical win ratio, calculated as the win ratio for the attacker in $100$ evaluation games.

The top row of Fig. \ref{fig:best_response_learning} demonstrates that the attacker quickly adjusts its policy to increase its win ratio. Specifically, we can observe that the policies converge after about $500$ iterations. Furthermore, we can see that our method, PPO-AR, achieves a higher win ratio than the two baselines in all three scenarios. It can also be seen that among the baselines, PPO outperforms REINFORCE.

Looking at the bottom row of Fig. \ref{fig:best_response_learning}, we observe, similar to the top row, that the defender's policy converges quickly, although the results tend to have a higher variance. Moreover, we see that our proposed method, PPO-AR, outperforms the baselines and achieves the lowest average win ratio of the attacker for all three scenarios. We also observe that PPO outperforms REINFORCE.

That PPO-AR outperforms the baselines indicates that the reduced action space that results from the autoregressive policy representation simplifies exploration, enabling the agents to discover more effective policies. Moreover, although we expected that PPO would do better than REINFORCE, as it uses a more theoretically justified policy update \cite{ppo}, it is surprising that the difference is so substantial in the results; we anticipate that it could be due to wrong setting of hyperparameters.

In the three columns of Fig. \ref{fig:best_response_learning} we see that the similarities and the differences among the three scenarios (see Section \ref{sec:scenarios}) are visible in the results. Specifically, that the win ratio of the attacker is lower in scenario $1$ and $2$ than in scenario $3$ reflects the fact that scenario $1$ and $2$ have strong initial defenses, whereas scenario $3$ has weak initial defenses. Similarly, that scenario $1$ and scenario $2$ have alike defenses can be seen on the similar curves in the two leftmost columns of Fig. \ref{fig:best_response_learning}. Moreover, the weak initial defense of scenario $3$ can be seen in the high variance of our method in scenario $3$ (Fig. \ref{fig:best_response_learning}, bottom row, right column). As all nodes are equally vulnerable in scenario $3$, the static attacker can select any attack path. Consequently, the defender learns a defense policy that predicts a diverse set of attacks, leading to a high variance.

\begin{figure*}
  \centering
    \scalebox{0.45}{
      \includegraphics{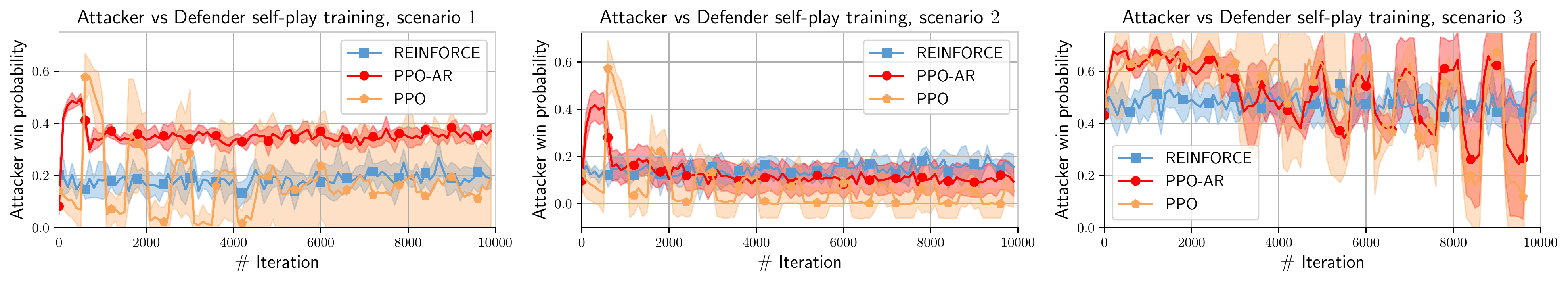}
    }
    \caption{Attacker win ratio against the number of training iterations; the three sub-graphs show the results from training the attacker and the defender in self-play for the three scenarios; the curve labeled PPO-AR shows the mean values of our proposed method; the results are averages over five training runs with different random seeds; the shaded regions show the standard deviation.}
    \label{fig:equlibrium_learning}
  \end{figure*}

\subsubsection{Inspection of Learned Policies}
We find that the learned \textit{attack} policies are deterministic and generally consist of two steps: (1) perform reconnaissance to collect information about neighboring nodes; and (2), exploit identified vulnerabilities (Fig. \ref{fig:idsgame_strategy}). Furthermore, we find that the learned \textit{defense} policies also are mostly deterministic and consist of hardening the critical node $N_{data}$, and patching identified vulnerabilities across all nodes. Finally, we also find that the defender learns to use the predictable attack pattern to its advantage and defends where the attacker is likely to attack next.
\subsection{Learning Policies Against a Dynamic Opponent}\label{sec:dnmp}
In this subsection we study the setting where the attacker and the defender learn simultaneously. Our goal is to investigate whether this setting can lead to stable and effective policies.
\subsubsection{Simulation Setup}
To evaluate our method, we use the same setup as in Section \ref{sec:brpl} and run $10,000$ training iterations, each with five different random seeds. After each iteration, \textit{both} agents update their policies. Hence, this setup is different from Section \ref{sec:brpl}, where one agent is static and only one agent updates its policy after each iteration. That both agents update their policies means that, in each iteration, an agent plays against a different opponent policy, making the environment non-stationary. Consequently, the main challenge in this setting is to get the agents' policies to converge. To improve convergence, we use the opponent pool technique described in Section \ref{sec:opponent_pool}. The results are shown in Fig. \ref{fig:equlibrium_learning}.
\subsubsection{Analysis of the Results in Fig. \ref{fig:equlibrium_learning}}
We observe in Fig. \ref{fig:equlibrium_learning} that policies sometimes converge and sometimes oscillate across iterations. We also see that this behavior is dependent on the particular scenario. For instance, we find that scenario $2$, which represents an infrastructure with good defenses (see Section \ref{sec:scenarios}), exhibits stable behavior of the policies. In contrast, scenario $3$, which captures an infrastructure with weak defenses, exhibits oscillation of the policies.

Looking at our proposed method, PPO-AR, we can see that it stabilizes in scenario $1$ and $2$ after some $1,000$ iterations. However, in scenario $3$ our method oscillates with increasing amplitude. The oscillation indicates that the agent reacts to the changed policy of its opponent and overfits. Similar behavior of self-play has been observed in related works \cite{self_play_cyclic, dota_openai_1}. Although the opponent pool technique has been introduced to mitigate the oscillation, it is clear from the results that this is insufficient in some scenarios.
\subsubsection{Inspection of Learned Policies}\label{sec:strategies_discussion}
When inspecting the learned policies, we find that they are fully stochastic. This contrasts with the policies inspected in Section \ref{sec:brpl}, that we found to be deterministic. By fully stochastic we mean that in a specific state of the game, the policy suggests several actions, each with significant probability. This result indicates that although deterministic policies are effective against static opponents, they are ineffective when facing adaptive opponents that can exploit the deterministic policy \cite{bowling_marl_1,heinrich_1}. For example, an adaptive attacker can learn to circumvent a deterministic defense policy. Finally, apart from being stochastic, the converged policies are similar to strategies of humans. In particular, we find that our method discovered stochastic defense policies that combine threat identification with attack forecasting and stochastic attack policies that combine reconnaissance with targeted exploits.

\subsection{Discussion of the Results}
The results demonstrate that, for a simple infrastructure, our method outperforms two baselines and discovers effective strategies for intrusion prevention without prior knowledge. When one agent is static in self-play, convergence is stable (Fig. \ref{fig:best_response_learning}) and the agents converge to deterministic policies that are effective against the static opponents. Conversely, when both agents learn simultaneously in self-play, convergence depends on the scenario (Fig. \ref{fig:equlibrium_learning}) and the agents converge to stochastic policies that are more general.

When inspecting the learned policies, we find that the policies reflect common-sense knowledge and are similar to strategies of humans. Most notably, these policies were discovered using no prior knowledge. This indicates that self-play works as a defense hardening technique. As the defender improves its defense in self-play, the attacker learns increasingly sophisticated attacks to circumvent the defense, leading to an artificial arms-race where both agents improve their policies over time.

In this paper, we have studied a relatively small and simple infrastructure (Fig. \ref{fig:model_graph}) and the application of our method to more complex\textemdash and more realistic\textemdash infrastructures is left for future work. We expect that by adding more complexity to the model, our method will be able to discover increasingly sophisticated strategies. However, we also expect that training times and computational requirements will increase.

Finally, the results also show that convergence of policies in self-play with two adaptive agents remains a challenge. In our results, we observe that the stability of policy convergence differs between the three scenarios (Fig. \ref{fig:equlibrium_learning}), indicating that the characteristic of the environment is an important factor for convergence in self-play. We also note that all policies that did converge in self-play with two adaptive agents were stochastic, which suggests that using methods that can represent stochastic policies is important for policy convergence in self-play.

\section{Related Work}
Our research extends prior work on reinforcement learning applied to network security, game-learning programs, and game theoretic modeling of security.

\subsection{Reinforcement Learning in Network Security}
The research on reinforcement learning applied to network security is in its infancy. Some early works are \cite{elderman,gt_learning_monitoring,honeypot_game,ridley_ml_defense}. For a literature review of deep reinforcement learning in network security, see \cite{deep_rl_cyber_sec}.

The prior work that most resembles ours are \cite{elderman}, and \cite{ridley_ml_defense}. Both of them experiment with abstract models for network intrusion and defense. Our work can be seen as a direct extension of these works as we use a similar model for our study. Our work in this paper differs in the following ways: (1) we propose an extension of the models in \cite{elderman,ridley_ml_defense} to include reconnaissance; (2) we analyze learned security strategies in self-play; (3) we propose a reinforcement learning method that uses function approximation, an opponent pool, and an autoregressive policy representation; and (4), we extend prior work to consider state-of-the-art algorithms (e.g. PPO \cite{ppo}).

\subsection{Game-Learning Programs}
Games have been studied since the inception of artificial intelligence and machine learning. An early example is Samuel's Checkers player \cite{samuel_1}. Other notable game-learning programs are Deep Blue \cite{deep_blue} and TD-Gammon \cite{td_gammon}. Recent developments include AlphaGo \cite{deepmind_2}, OpenAI Five \cite{dota_openai_1}, DeepStack \cite{deepstack}, and Libratus \cite{libratus}.

Although we take inspiration from these works, our study differs by focusing on security strategies as opposed to pure game strategies. Moreover, even though our study focuses on self-play training, we also take inspiration from prior work that have investigated learning theory in games, in particular Minimax-Q \cite{littman_marl_1}, Nash-Q \cite{nash_q_l}, Neural Fictitious Self-Play \cite{heinrich_1}, and the seminal book by Fudenberg and Levine \cite{fudenberg_levine}.
\subsection{Game Theoretic Modeling in Cyber Security}
Several examples of game theoretic security models can be found in the literature \cite{game_t_sec_survey, flipit, nework_security_alpcan}. For example, FlipIt \cite{flipit} is a security game that models Advanced Persistent Threats (APTs). Other notable works include the book by Alpcan and Basar \cite{nework_security_alpcan} and the survey by Manshaei et al. \cite{game_t_sec_survey}. We consider the related work on game theoretic methods to be complementary to the reinforcement learning methods studied in this paper.

\section{Conclusion and Future Work}
We have proposed a method to automatically find security strategies for the use case of intrusion prevention. The method consists of modeling the interaction between an attacker and a defender as a Markov game and letting attack and defense strategies evolve through reinforcement learning and self-play without human intervention. We have also addressed known challenges of reinforcement learning in this domain and presented an approach that uses function approximation, an opponent pool, and an autoregressive policy representation.

Using a simple infrastructure configuration, we have demonstrated that effective security strategies can emerge from self-play. Inspection of the converged policies show that the emerged policies reflect common-sense knowledge and are similar to strategies of humans (e.g. Fig. \ref{fig:idsgame_strategy}). Moreover, through evaluations, we have demonstrated that our method is superior to two baseline methods (Fig. \ref{fig:best_response_learning}), but that the problem of oscillating policies in self-play remains a challenge (Fig. \ref{fig:equlibrium_learning}). In particular, we have found a scenario where our method does not lead to converging but oscillating policies.

In future work we plan to further study the observed policy oscillations and explore techniques for mitigation. We also plan to increase the scale of the intrusion prevention use case and model more realistic infrastructures, e.g. a campus network. Finally, we will consider two extensions to our model. First, we plan to enrich the model's action space to cover additional use-cases, e.g. Advanced Persistent Threats (APTs). Second, we plan to lower the level of abstraction of the model to examine if our method can discover practically useful security strategies, i.e. strategies related to specific network protocols and software.

\section{Acknowledgments}
This research is supported in part by the Swedish armed forces and was conducted at KTH Center for Cyber Defense and Information Security (CDIS). The authors would like to thank Pontus Johnson, Professor at KTH, for useful input to this research, and Rodolfo Villaça and Forough Shahab Samani for their constructive remarks.

\appendix
\subsection{Hyperparameters}\label{appendix:hyperparameters}
\begin{table}[H]
\centering
\begin{tabular}{ll} \toprule
    {\textit{Parameter}} & {\textit{Value}} \\ \midrule
  Discount factor $\gamma$ & $0.999$  \\
  Learning rate $\alpha$ & $0.0001$ \\
  \# Shared layers & $2$ \\
  \# Hidden neurons & $128$ \\
  \# Layers for value head & $1$ \\
  \# Layers for policy head & $1$ \\
  Batch size & $2000$ \\
  Entropy coefficient & $0.001$ \\
  GAE $\lambda$  & $0.95$ \\
  Clipping range $\epsilon$ & $0.2$ \\
  Optimizer & Adam \cite{adam_opt} \\
  Opponent pool max size & 100000 \\
  Opponent pool increment iterations & 50 \\
  Opponent pool sampĺe $p$ & $0.5$ \\
  \bottomrule\\
\end{tabular}
\caption{Hyperparameters.}\label{tab:hyperparams}
\end{table}

\bibliographystyle{IEEEtran}
\bibliography{references}

\end{document}